\documentclass[10pt,twocolumn,letterpaper]{article}

\usepackage{wacv}
\usepackage{times}
\usepackage{epsfig}
\usepackage{graphicx}
\usepackage{amsmath}
\usepackage{amssymb}
\usepackage{booktabs}
\usepackage{tabularx}
\usepackage{multirow}
\usepackage{booktabs}
\usepackage{algorithm}
\usepackage{algpseudocode}
\usepackage{setspace}
\usepackage{amsfonts,amssymb}
\usepackage{pifont}
\usepackage{adjustbox}
\usepackage{enumitem}
\setlist[itemize]{noitemsep,topsep=3pt,leftmargin=*}
\usepackage{bm}
\usepackage{subfigure}
\usepackage{makecell}
\usepackage[dvipsnames]{xcolor}
\usepackage[accsupp]{axessibility}
\usepackage{marvosym}
\usepackage{ifsym}

\def\name{FORMULA}

\def\wacvPaperID{***} 

\wacvalgorithmstrack   

\wacvfinalcopy 

\ifwacvfinal
\usepackage[breaklinks=true,bookmarks=false]{hyperref}
\else
\usepackage[pagebackref=true,breaklinks=true,colorlinks,bookmarks=false]{hyperref}
\fi

\begin{document}

\title{Foreground Guidance and Multi-Layer Feature Fusion for Unsupervised Object Discovery with Transformers}


\author{Zhiwei Lin$^{*}$
    \and 
    Zengyu Yang$^{*\dagger}$
    \and Yongtao Wang$^{\textrm{\Letter}}$
    \and Wangxuan Institute of Computer Technology, Peking University \\
	{\tt\small zwlin@pku.edu.cn, yzyysj@gmail.com, wyt@pku.edu.cn}
}

\maketitle
\thispagestyle{empty}

\begin{abstract}
Unsupervised object discovery (UOD) has recently shown encouraging progress with the adoption of pre-trained Transformer features.
However, current methods based on Transformers mainly focus on designing the localization head (\eg, seed selection-expansion and normalized cut) and overlook the importance of improving Transformer features. 
In this work, we handle UOD task from the perspective of feature enhancement and propose \textbf{FOR}eground guidance and \textbf{MU}lti-\textbf{LA}yer feature fusion for unsupervised object discovery, dubbed \name.
Firstly, we present a foreground guidance strategy with an off-the-shelf UOD detector to highlight the foreground regions on the feature maps and then refine object locations in an iterative fashion.
Moreover, to solve the scale variation issues in object detection, we design a multi-layer feature fusion module that aggregates features responding to objects at different scales.
The experiments on VOC07, VOC12, and COCO\_20k show that the proposed \name~achieves new state-of-the-art results on unsupervised object discovery. 
The code will be released at \href{https://github.com/VDIGPKU/FORMULA}{https://github.com/VDIGPKU/FORMULA}.
\end{abstract}

\begin{figure}[ht!]
	\centering
	\subfigure[LOST]{
		\includegraphics[height=2.4cm]{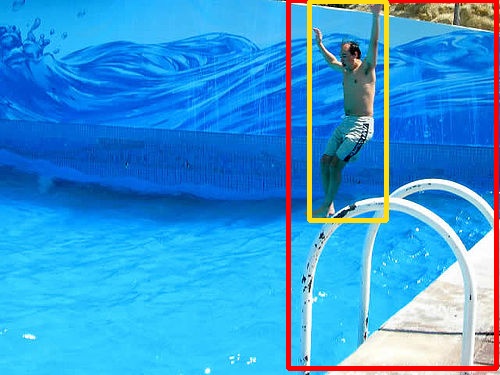}\label{fig_1a}
	}
	\hspace{1mm}
	\subfigure[TokenCut]{
		\includegraphics[height=2.4cm]{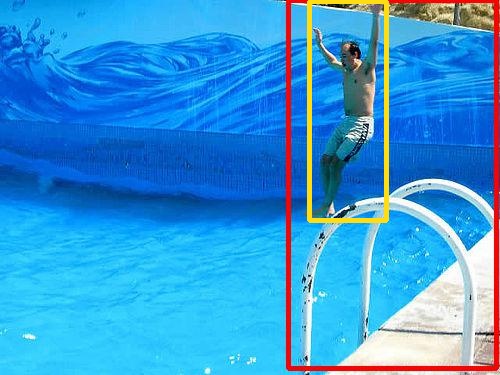}\label{fig_1b}
	}
	\hspace{1mm}
	\subfigure[\name-L (ours)]{
		\includegraphics[height=2.4cm]{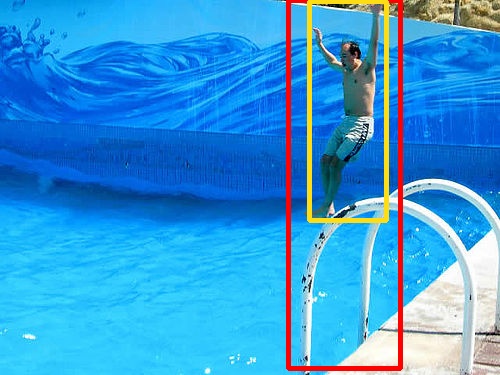}\label{fig_1c}
	}
	\hspace{1mm}
	\subfigure[\name-TC (ours)]{
		\includegraphics[height=2.4cm]{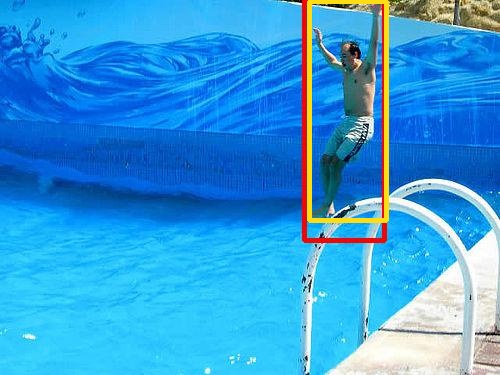}\label{fig_1d}
	}
	\caption{{\bf Example results of UOD on VOC12.} 
	In (a) and (b), we show results obtained by LOST~\cite{DBLP:conf/bmvc/SimeoniPVRGBPMP21} and TokenCut~\cite{wang2022tokencut} respectively, which are two state-of-the-art UOD methods. The results of our method are presented in (c) and (d). Predictions are in \textcolor{red}{red}, and ground-truth boxes are in \textcolor{Dandelion}{yellow}. 
	We can find that the proposed \name~localizes object more accurately.
	Best viewed in color. 
	}
	\label{fig:fig1}
\end{figure}

\section{Introduction}
\footnotetext{$^*$Equal contribution. $^{\dagger}$As an intern at PKU. $^{\textrm{\Letter}}$Corresponding author.}
Object detection is one of the fundamental problems in computer vision, which serves a wide range of applications such as face recognition~\cite{DBLP:conf/cvpr/TaigmanYRW14}, pose estimation~\cite{DBLP:conf/iccv/YangQNY21}, and autonomous driving~\cite{DBLP:conf/ivs/TeichmannWZCU18}.
In recent years, significant success has been achieved in the field~\cite{DBLP:conf/nips/RenHGS15,DBLP:conf/cvpr/RedmonDGF16} thanks to the increasing amount of annotated training data. However, the labeling of large-scale datasets~\cite{DBLP:conf/eccv/LinMBHPRDZ14,DBLP:journals/ijcv/RussakovskyDSKS15} is rather costly. Although multiple techniques, including semi-supervised learning~\cite{DBLP:conf/nips/BerthelotCGPOR19}, weakly-supervised learning~\cite{Feng_2022_CVPR}, and self-supervised learning~\cite{gidaris2018unsupervised} have been proposed to alleviate this issue, manual labeling is still required.

Here we concentrate on a fully unsupervised task for object detection, named unsupervised object discovery.
Previous CNN-based methods~\cite{DBLP:conf/cvpr/ChoKSP15,DBLP:conf/cvpr/VoBCHLPP19,DBLP:conf/eccv/VoPP20,DBLP:journals/pr/WeiZWSZ19} leverage region proposal and localize objects by comparing the proposed bounding boxes of each image across a whole image collection.
However, these approaches are difficult to scale to large datasets due to the quadratic complexity brought by the comparison process~\cite{DBLP:conf/nips/VoSSPP21}. Recently, DINO~\cite{DBLP:conf/iccv/CaronTMJMBJ21} has found that the attention maps of a Vision Transformer (ViT)~\cite{DBLP:conf/iclr/DosovitskiyB0WZ21} pre-trained with self-supervised learning reveal salient foreground regions. Motivated by DINO, LOST~\cite{DBLP:conf/bmvc/SimeoniPVRGBPMP21} and TokenCut~\cite{wang2022tokencut} are proposed to discover objects by leveraging the high-quality ViT features. Both methods first construct an undirected graph using the similarity of patch-wise features from the last layer of the ViT. Then, a two-step seed-selection-expansion strategy and Normalized Cut~\cite{DBLP:journals/pami/ShiM00} are adopted respectively to segment the foreground objects. While both approaches have achieved superior results over previous state-of-the-art~\cite{DBLP:conf/eccv/VoPP20,DBLP:conf/nips/VoSSPP21}, we have found that they mainly focus on the construction of the localization head and overlook the potential of improving the ViT features.

In this paper, we propose a simple but effective feature enhancement method for existing ViT-based UOD frameworks, named \name.
Our method consists of two parts, \textit{i.e.}, the foreground guidance module and the multi-layer feature fusion module. For the foreground guidance module, we utilize the object mask predicted by an off-the-shelf UOD detector to highlight the foreground object region and then refine the object location through an iterative process.
Specifically, we first generate an object mask from the original ViT feature map using an existing UOD detector (\eg, LOST or TokenCut). 
Then, we construct a probability map with 2D Gaussian distribution from the mask, which roughly localizes the foreground objects. 
After that, we highlight the foreground area by applying the probability map to the original ViT feature map. 
Finally, the updated feature map is used for the UOD detector to obtain a refined object mask. The whole process can be iterated. In this way, we enhance the ViT feature map by introducing the foreground object information and suppressing background interference. Our method can localize objects much more accurately with only a few iterations.

Besides, we note that LOST and TokenCut only use the feature map from the last layer of ViT. However, the scale of objects in non-object-centric images, like those in COCO~\cite{DBLP:conf/eccv/LinMBHPRDZ14}, can vary greatly. The feature from the last layer of a pre-trained ViT mainly captures the key areas for classification, which is usually at a larger scale. Thus, the performance on smaller objects is hurt. To address this issue, we propose the multi-layer feature fusion module. In detail, we simply merge the features from the last several layers through a weighted summation to aggregate multi-scale information for unsupervised object discovery.

Our main contributions can be summarized as follows:
\begin{itemize}
	\item We introduce foreground guidance from the object mask predicted by an existing UOD detector to the original ViT feature map and propose an iterative process to refine the object location.\\

	\item We further design a multi-layer feature fusion module to address the scale variation issues in object detection, releasing the potential of ViT feature representation for object discovery.\\
	
	\item The proposed method can be easily incorporated into any existing UOD methods based on ViT and achieves new state-of-the-art results on the popular PASCAL VOC and COCO benchmarks.
\end{itemize}

\section{Related Work}
{\bf Self-supervised learning.} 
Learning powerful feature representations in a self-supervised manner that dispenses with human annotation has made great progress recently. This is performed by defining a pretext task that provides surrogate supervision for feature learning~\cite{DBLP:conf/eccv/NorooziF16,DBLP:conf/eccv/ZhangIE16,gidaris2018unsupervised}. Despite no labels, models trained with self-supervision have outperformed their supervised counterparts on several downstream tasks, such as image classification~\cite{gidaris2018unsupervised,DBLP:conf/cvpr/WuXYL18,DBLP:conf/icml/ChenK0H20,DBLP:conf/cvpr/He0WXG20,DBLP:journals/corr/abs-2003-04297,DBLP:conf/nips/CaronMMGBJ20,DBLP:conf/nips/GrillSATRBDPGAP20,DBLP:conf/cvpr/ChenH21,DBLP:conf/iccv/CaronTMJMBJ21,wang2021self,DBLP:conf/iccv/ChenXH21,https://doi.org/10.48550/arxiv.2111.06377} and object detection~\cite{DBLP:conf/nips/WeiGWHL21,DBLP:conf/iccv/ChenHXLY21}. While~\cite{DBLP:conf/cvpr/WuXYL18,DBLP:conf/icml/ChenK0H20,DBLP:conf/cvpr/He0WXG20,DBLP:conf/nips/CaronMMGBJ20,DBLP:conf/nips/GrillSATRBDPGAP20,DBLP:conf/cvpr/ChenH21} have adopted CNN as pre-training backbones, recent works~\cite{DBLP:conf/iccv/CaronTMJMBJ21,wang2021self,DBLP:conf/iccv/ChenXH21,https://doi.org/10.48550/arxiv.2111.06377,DBLP:conf/nips/LiCYLZZDWZTW21,li2021esvit} have explored Transformers~\cite{DBLP:conf/nips/VaswaniSPUJGKP17} for self-supervised visual learning, demonstrating their superiority over traditional CNN. Our work utilizes the strong localization capability of self-supervised ViT for unsupervised object discovery.

\noindent
{\bf Unsupervised object discovery.} The goal of unsupervised object discovery is to localize objects without any supervision. Early works generally rely on image features encoded by a CNN~\cite{ijcai2018-104,DBLP:conf/eccv/VoPP20,DBLP:journals/pr/WeiZWSZ19,DBLP:conf/nips/VoSSPP21,DBLP:journals/corr/abs-1906-05857}.
These methods need to compare the extracted features of each image across those of every other image in the dataset, leading to quadratic computational overhead~\cite{DBLP:conf/nips/VoSSPP21}. 
Besides, the dependence on the inter-image similarity results in these methods being unable to run on a single image. 
Recently, LOST~\cite{DBLP:conf/bmvc/SimeoniPVRGBPMP21} and TokenCut~\cite{wang2022tokencut} have been proposed to address these issues by leveraging the high-quality feature representation generated from a self-supervised ViT. Their motivation is that the attention map extracted from the last layer of a pre-trained ViT contains explicit information about the foreground objects. 
Specifically, both methods first propose the construction of an intra-image similarity graph using features extracted from the ViT backbone~\cite{DBLP:conf/iccv/CaronTMJMBJ21,DBLP:conf/iclr/DosovitskiyB0WZ21}. A heuristic seed-selection-expansion strategy and Normalized Cut~\cite{DBLP:journals/pami/ShiM00} are then adopted respectively by the two methods to segment a single foreground object in an image.

Although achieving excellent performance, these methods mainly concentrate on localization design and fail to further improve ViT features for unsupervised object discovery. Instead, our work starts from the perspective of feature enhancement. Concretely, we introduce the foreground guidance to highlight the foreground regions on ViT features and propose a multi-layer feature fusion module to aggregate multi-scale features.

\begin{figure*}[!ht]
	\centering
	\includegraphics[width=16cm]{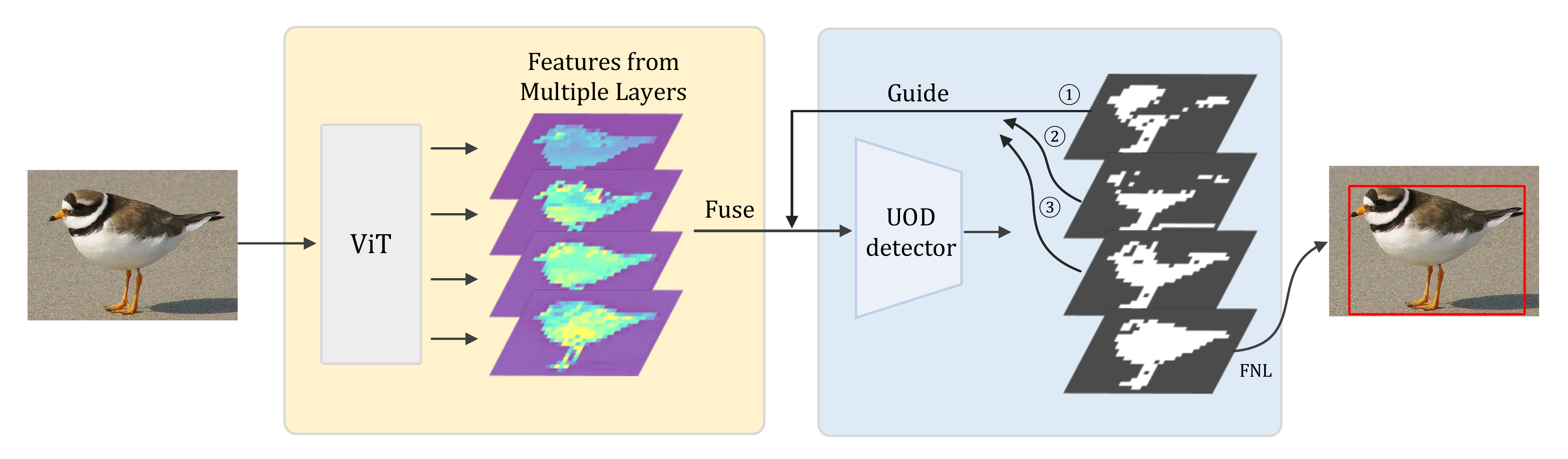}
	\caption{{\bf The pipeline of \name.} To enhance ViT features for unsupervised object discovery, we extract multi-layer features to aggregate information from different scales and introduce foreground guidance from the predicted segmentation to the input of the UOD detector.
	}
	\label{fig:pipeline}
\end{figure*}

\noindent
{\bf Multi-layer Feature Representations.}
One of the main challenges in object detection is to effectively represent features on different scales. Extensive works have been proposed over the years to deal with the multi-scale problem with multi-layer features. These methods leverage the pyramidal features of CNN to compute a multi-scale feature representation.~\cite{DBLP:conf/cvpr/LongSD15,DBLP:conf/cvpr/HariharanAGM15,DBLP:conf/cvpr/KongYCS16,DBLP:conf/cvpr/BellZBG16,DBLP:conf/miccai/RonnebergerFB15,DBLP:conf/eccv/GhiasiF16,DBLP:conf/cvpr/LinDGHHB17} combine low-resolution and high-resolution features with upsampling or lateral connections to aggregate semantic information from all levels.~\cite{DBLP:conf/eccv/LiuAESRFB16,DBLP:conf/eccv/CaiFFV16,DBLP:conf/cvpr/LinDGHHB17} make predictions at different scales from different layers and use post-processing to filter the final predictions. In addition to CNN, several works have recently exploited the multi-layer representation for Transformer networks.~\cite{DBLP:conf/bmvc/WangY021} aggregates the class tokens from each Transformer layer to gather the local, low-level, and middle-level information that is crucial for fine-grained visual categorization. In~\cite{https://doi.org/10.48550/arxiv.2207.14467}, the authors divide the multi-layer representations of the Transformer’s encoder and decoder into different groups and then fuse these group features to fully leverage the low-level and high-level features for neural machine translation. 

These works inspire us to explore the multi-layer features of ViT for better object localization. Instead of designing complicated fusion modules, we propose a simple and efficient fusion method that sums the feature from each layer of the Transformer with different weights.

\section{Approach}
In this section, we introduce our method for unsupervised object discovery, \textit{i.e.}, \name. The overall pipeline of \name~is presented in Fig.~\ref{fig:pipeline}. Firstly, we briefly review Vision Transformers and their previous applications in UOD as preliminary knowledge. Then, we describe the two modules of \name, namely foreground guidance and multi-layer feature fusion.

\subsection{Preliminary}
Vision Transformers~\cite{DBLP:conf/iclr/DosovitskiyB0WZ21} receive a sequence of image patches and use stacked multi-head self-attention blocks to extract feature maps from images. It divides an input image of $H\times W$ into a sequence of $N=HW/P^{2}$ patches of fixed resolution $P\times P$. Patch embeddings are then formed by mapping the flattened patches to a $D$-dimensional latent space with a trainable linear projection. An extra learnable \verb'[CLS]' token is attached to the patch embeddings and position embeddings are added to form the standard transformer input in $\mathbb{R}^{(N+1)\times D}$.

DINO~\cite{DBLP:conf/iccv/CaronTMJMBJ21} has shown that the attention map extracted from the last layer of a self-supervised ViT indicates prominent foreground areas. Following this observation, LOST
and TokenCut
propose to localize objects using the key features $k\in \mathbb{R}^{N\times D}$ from the last layer in two steps. First, an intermediate feature map $F_{int}\in \mathbb{R}^{h\times w}$ is constructed from the inter-patch similarity graph, where $h=H/P$ and $w=W/P$. 
Specifically, for LOST, it is the map of inverse degrees; for TokenCut, it is the second smallest eigenvector of the graph. 
Second, a object mask $m\in \{0, 1\}^{h\times w}$ is generated from $F_{int}$ to segment the foreground object.

\begin{figure*}[htbp]
	\centering
	\includegraphics[width=14cm]{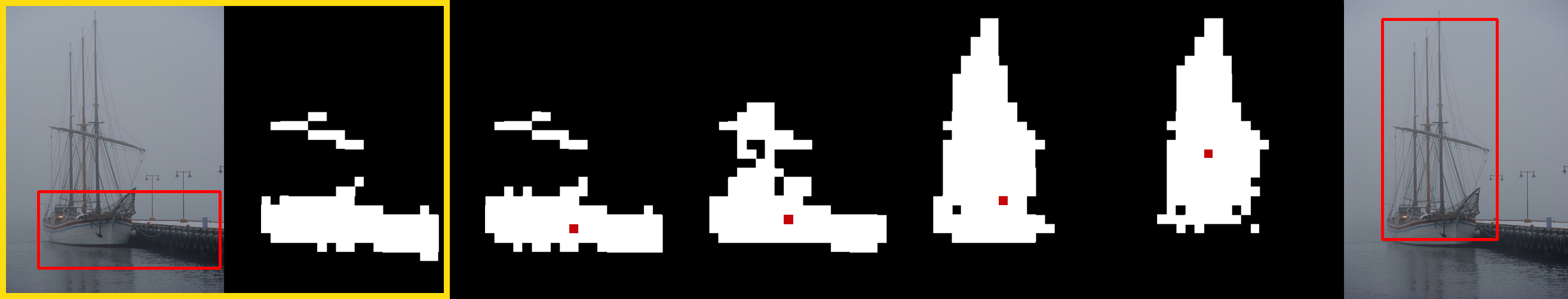}
	\caption{{\bf Illustration of the iteration process.} The initial prediction made by LOST~\cite{DBLP:conf/bmvc/SimeoniPVRGBPMP21} (framed in \textcolor{Dandelion}{yellow}) and the final prediction after four iterations (right). The red dot in each iteration is the object center calculated from the previous segmentation.}
	\label{fig:iteration}
\end{figure*}

\subsection{Foreground Guidance with Self-iteration}
In the foreground guidance module, the predicted object mask is treated as the foreground guidance to highlight the foreground region and guide the segmentation process. Specifically, given an existing unsupervised object detector $\mathcal{D}$ and an intermediate feature map $F_{int}$ extracted from the pre-trained ViT, the binary foreground object mask $m\in \{0, 1\}^{h\times w}$ of the object can be generated as follows:

\begin{equation}
	\label{eq:eq1}
	m=\mathcal{D}(F_{int}).
\end{equation}

\noindent
Here, $\mathcal{D}$ could be any ViT-based object discovery methods, \textit{e.g.}, LOST and TokenCut. Moreover, the value of $m(\bm{x}_i)$ equals to 1 if the corresponding patch $i$ with coordinates $\bm{x}_i$ is predicted to belong to the foreground object. With the foreground mask $m$, the approximate coordinates of the object center $\mathcal{O}$ can be calculated by
\begin{equation}
	\label{eq:eq2}
	\texttt{}
	\bm{x}_{\mathcal{O}}=\frac{1}{\sum_{i=1}^{h\times w} m(\bm{x}_i)}\sum_{i=1}^{h\times w}m(\bm{x}_i)\bm{x}_i.
\end{equation}

\noindent
Then, we construct a probability map $P\in \mathbb{R}^{h\times w}$ using the 2D Gaussian distribution function $g$:

\begin{equation}
	\label{eq:eq3}
	P(i)=g(i|\bm{x}_{\mathcal{O}},\sigma^{2})=\frac{1}{2\pi \sigma^{2}}e^{-\frac{\|\bm{x}_{i}-\bm{x}_{\mathcal{O}}\|^{2}}{2\sigma^{2}}},
\end{equation}

\noindent
where $\sigma$ is a hyper-parameter. Intuitively, the value of $P$ indicates the regions in an image that are likely to belong to an object. The probability map $P$ can be viewed as the foreground guidance for object localization, guiding the detector $\mathcal{D}$ to refine the final prediction in each iteration step. 
Specifically, we achieve this by applying the Hadamard product to $P$ and $F_{int}$:

\begin{equation}
	\label{eq:eq4}
	\widetilde{F_{int}}=F_{int}\circ P.
\end{equation}

\noindent
The new feature map $\widetilde{F_{int}}$ can be interpreted as a re-weighting of $F_{int}$ and the foreground part of $F_{int}$ is emphasized. Thus, during unsupervised object discovery, the detector will focus more on the foreground part instead of the whole image.

Finally, we can generate a new object mask through Eq. \ref{eq:eq1} from $\widetilde{F_{int}}$ and iterate the whole procedure until convergence, \textit{i.e.}, the distance between the centers in two consecutive iterations is smaller than $\tau$. The overall process is presented in Algorithm \ref{alg:alg1}. An example of the iteration process is illustrated in Fig. \ref{fig:iteration}. It is worth noting that the efficiency bottleneck of ViT-based UOD methods lies in the process of feature extraction rather than localization. Thus, our iteration module only brings marginal extra computational overhead. More details can be found in \ref{sec:ablations}.

\begin{algorithm}[htbp]
	\caption{Foreground Guidance with Self-iteration}
	\label{alg:alg1}
	\setstretch{1.08}
	\begin{algorithmic}
		\State {\bf Input:} Unsupervised Object Detector $\mathcal{D}$, Intermediate Feature Map $F_{int}$, Standard Deviation $\sigma$ of Gaussian Distribution $g$.
		\State {\bf Initialize:} $m=\mathcal{D}(F_{int})$, $\bm{x}_{\mathcal{O}}=0$, $\widehat{\bm{x}_{\mathcal{O}}}=\infty$.
		\While{$\|\widehat{\bm{x}_{\mathcal{O}}}-\bm{x}_{\mathcal{O}}\|^{2}\geqslant\tau$}
		\State $\widehat{\bm{x}_{\mathcal{O}}}=\bm{x}_\mathcal{O}$
		\State $\bm{x}_\mathcal{O}=Eq.\ \ref{eq:eq2}(m)$
		\State $P=g(\bm{x}_\mathcal{O},\sigma^{2})$
		\State $\widetilde{F_{int}}=F_{int}\circ P$
		\State $m=\mathcal{D}(\widetilde{F_{int}})$
		\EndWhile
		\State {\bf Output:} $m_{ref}=m$.
	\end{algorithmic}
\end{algorithm}

\subsection{Multi-layer Feature Fusion}
Different layers of Transformers encode features at different scales.
Deeper layers tend to gather global and semantic information and focus on the discriminative parts of objects. 
Consequently, the activated areas in their feature maps are smaller (Fig. \ref{fig_5b}, \ref{fig_5c}). 
In contrast, shallower layers focus on local information
and thus the activated areas in their feature maps are broader
(Fig. \ref{fig_5d}, \ref{fig_5e}). 
However, the scale of the objects in non-object-centric images can vary greatly. Only using the feature from one layer is insufficient to deal with the scale variation problem. To address this issue, we propose the multi-layer feature fusion module to aggregate information from various scales.

In detail, from each multi-head attention layer $l$ of a ViT, we can extract the key feature $k_{l}\in \mathbb{R}^{(N+1)\times D}$. Then, we drop the \verb'[CLS]' token to be consistent with LOST and TokenCut. The aggregated feature for unsupervised object discovery is obtained by a weighted summation of the key features from all layers:

\begin{equation}
	\label{eq:eq5}
	f=\sum_{l=1}^{L}\alpha_{l}k_{l},
\end{equation}

\noindent
where $\alpha_{l}$ is the weight of layer $l$. The features from different layers of ViT contain information of objects at various scales. The aggregated feature incorporates more comprehensive object information at different scales to better localize objects.

\section{Experiments}
\label{sec:exp}
In this section, we conduct extensive experiments on various datasets to demonstrate the effectiveness of our method. \name~achieves new state-of-the-art results in unsupervised object discovery tasks. In addition, we conduct ablation studies to discuss the effect of foreground guidance and multi-layer feature fusion modules.

\begin{table*}[!h]
	\begin{center}
		\begin{tabular}{lcccc}
			\toprule
			Method & Backbone & VOC07$(\uparrow)$ & VOC12$(\uparrow)$ & COCO\_20k$(\uparrow)$ \\
			\midrule
			Selective Search~\cite{DBLP:journals/ijcv/UijlingsSGS13} & - & 18.8 & 20.9 & 16.0 \\
			EdgeBoxes~\cite{10.1007/978-3-319-10602-1_26} & - & 31.1 & 31.6 & 28.8 \\
			Kim \etal~\cite{NIPS2009_a87ff679} & - & 43.9 & 46.4 & 35.1 \\
			Zhange \etal~\cite{DBLP:journals/corr/abs-1902-09968} & VGG16~\cite{DBLP:journals/corr/SimonyanZ14a} & 46.2 & 50.5 & 34.8 \\
			DDT+~\cite{DBLP:journals/pr/WeiZWSZ19} & VGG19~\cite{DBLP:journals/corr/SimonyanZ14a} & 50.2 & 53.1 & 38.2 \\
			rOSD~\cite{DBLP:conf/eccv/VoPP20} & VGG16\&19~\cite{DBLP:journals/corr/SimonyanZ14a} & 54.5 & 55.3 & 48.5 \\
			LOD~\cite{DBLP:conf/nips/VoSSPP21} & VGG16~\cite{DBLP:journals/corr/SimonyanZ14a} & 53.6 & 55.1 & 48.5 \\
			DINO-seg~\cite{DBLP:conf/iccv/CaronTMJMBJ21,DBLP:conf/bmvc/SimeoniPVRGBPMP21} & ViT-S/16~\cite{DBLP:conf/iccv/CaronTMJMBJ21,DBLP:conf/iclr/DosovitskiyB0WZ21} & 45.8 & 46.2 & 42.1 \\
			LOST~\cite{DBLP:conf/bmvc/SimeoniPVRGBPMP21} & ViT-S/16~\cite{DBLP:conf/iccv/CaronTMJMBJ21,DBLP:conf/iclr/DosovitskiyB0WZ21} & 61.9 & 64.0 & 50.7 \\
			TokenCut~\cite{wang2022tokencut} & ViT-B/16~\cite{DBLP:conf/iccv/CaronTMJMBJ21,DBLP:conf/iclr/DosovitskiyB0WZ21} & 68.8 & 72.4 & 59.0 \\
			\midrule
			\name-L & ViT-S/16~\cite{DBLP:conf/iccv/CaronTMJMBJ21,DBLP:conf/iclr/DosovitskiyB0WZ21} & 64.4 & 67.7 & 54.0 \\
			{\bf \name-TC} & ViT-B/16~\cite{DBLP:conf/iccv/CaronTMJMBJ21,DBLP:conf/iclr/DosovitskiyB0WZ21} & {\bf 69.4} & {\bf 73.2} & {\bf 59.7} \\
			\midrule
			\midrule
			LOD + CAD~\cite{DBLP:conf/bmvc/SimeoniPVRGBPMP21} & VGG16~\cite{DBLP:journals/corr/SimonyanZ14a} & 56.3 & 61.6 & 52.7 \\
			rOSD + CAD~\cite{DBLP:conf/bmvc/SimeoniPVRGBPMP21} & VGG16\&19~\cite{DBLP:journals/corr/SimonyanZ14a} & 58.3 & 62.3 & 53.0 \\
			LOST + CAD~\cite{DBLP:conf/bmvc/SimeoniPVRGBPMP21} & ViT-S/16~\cite{DBLP:conf/iccv/CaronTMJMBJ21,DBLP:conf/iclr/DosovitskiyB0WZ21} & 65.7 & 70.4 & 57.5 \\
			TokenCut + CAD$^\ast$~\cite{DBLP:conf/bmvc/SimeoniPVRGBPMP21} & ViT-B/16~\cite{DBLP:conf/iccv/CaronTMJMBJ21,DBLP:conf/iclr/DosovitskiyB0WZ21} & 71.4 & 75.5 & 62.7 \\
			\midrule
			\name-L + CAD~\cite{DBLP:conf/bmvc/SimeoniPVRGBPMP21} & ViT-S/16~\cite{DBLP:conf/iccv/CaronTMJMBJ21,DBLP:conf/iclr/DosovitskiyB0WZ21} & 66.8 & 71.5 & 59.0 \\
			{\bf \name-TC + CAD}~\cite{DBLP:conf/bmvc/SimeoniPVRGBPMP21} & ViT-B/16~\cite{DBLP:conf/iccv/CaronTMJMBJ21,DBLP:conf/iclr/DosovitskiyB0WZ21} & {\bf 72.2} & {\bf 76.2} & {\bf 63.7} \\
			\bottomrule
		\end{tabular}
	\end{center}
	\caption{{\bf Unsupervised single-object discovery.} We compare \name~with current state-of-the-art unsupervised object discovery methods. We use DINO pre-trained ViT~\cite{DBLP:conf/iccv/CaronTMJMBJ21} as our backbone. $^{\ast}$ is our implementation. `-L/TC' means to adopt LOST/TokenCut as the unsupervised object detector $\mathcal{D}$. `+CAD' means to train a second-stage object detector using pseudo-labels generated through each method.}
	\label{tab:main_result}
\end{table*}

\begin{table*}[htbp]
	\Huge
	\begin{center}
		\resizebox{\textwidth}{!}{
			\begin{tabular}{l|c|cccccccccccccccccccc|c}
				\toprule
				Method & Supervis. & aero & bike & bird & boat & bottle & bus & car & cat & chair & cow & table & dog & horse & mbike & person & plant & sheep & sofa & train & tv & mean$(\uparrow)$\\
				\midrule
				WSDDN~\cite{DBLP:conf/cvpr/BilenV16} & weak & 39.4 & 50.1 & 31.5 & 16.3 & 12.6 & 64.5 & 42.8 & 42.6 & 10.1 & 35.7 & 24.9 & 38.2 & 34.4 & 55.6 & 9.4 & 14.7 & 30.2 & 40.7 & 54.7 & 46.9 & 34.8 \\
				PCL~\cite{DBLP:journals/pami/TangWBSBLY20} & weak & 54.4 & 9.0 & 39.3 & 19.2 & 15.7 & 62.9 & 64.4 & 30.0 & 25.1 & 52.5 & 44.4 & 19.6 & 39.3 & 67.7 & 17.8 & 22.9 & 46.6 & 57.5 & 58.6 & 63.0 & 43.5\\
				rOSD~\cite{DBLP:conf/eccv/VoPP20} + OD~\cite{DBLP:conf/bmvc/SimeoniPVRGBPMP21} & - & 38.8 & 44.7 & 25.2 & 15.8 & 0.0 & 52.9 & 45.4 & 38.9 & 0.0 & 16.6 & 24.4 & 43.3 & 57.2 & 51.6 & 8.2 & 0.7 & 0.0 & 9.1 & 65.8 & 9.4 & 27.4 \\
				LOST~\cite{DBLP:conf/bmvc/SimeoniPVRGBPMP21} + OD~\cite{DBLP:conf/bmvc/SimeoniPVRGBPMP21} & - & 57.4 & 0.0 & 40.0 & 19.3 & 0.0 & 53.4 & 41.2 & 72.2 & 0.2 & 24.0 & 28.1 & 55.0 & 57.2 & 25.0 & 8.3 & 1.1 & 0.9 & 21.0 & 61.4 & 5.6 & 28.6\\
				\midrule
				\name-L + OD~\cite{DBLP:conf/bmvc/SimeoniPVRGBPMP21} & - & 62.8 & 3.2 & 45.6 & 23.9 & 0.0 & 53.3 & 41.2 & 74.3 & 0.1 & 18.9 & 32.7 & 60.7 & 59.8 & 27.2 & 11.4 & 0.0 & 0.0 & 38.5 & 39.8 & 2.3 & 29.8\\
				{\bf \name-TC + OD}~\cite{DBLP:conf/bmvc/SimeoniPVRGBPMP21} & - & 61.4 & 19.2 & 49.4 & 26.1 & 0.0 & 60.5 & 46.6 & 78.7 & 0.4 & 21.2 & 31.8 & 73.7 & 55.2 & 15.8 & 12.2 & 0.0 & 0.0 & 41.5 & 51.5 & 8.9 & 32.7\\
				\bottomrule
		\end{tabular}}
	\end{center}
	\caption{\small{\bf Unsupervised object detection (OD).} We evaluate \name~on VOC07 {\tt test} using AP@0.5. All the `+OD' methods are trained on VOC07 {\tt trainval}.}
	\label{tab:od}
\end{table*}

\subsection{Experimental Settings}
\label{sec:exp_settings}
\noindent
{\bf Datasets.} Following LOST and TokenCut, we evaluate the proposed approach on three widely-adopted benchmarks for object discovery: VOC07~\cite{pascal-voc-2007}, VOC12~\cite{pascal-voc-2012}, and COCO~\cite{DBLP:conf/eccv/LinMBHPRDZ14}. Specifically, for VOC07 and VOC12, we use \verb'trainval' set to evaluate our method. For COCO, we only use a subset of the COCO2014 \verb'trainval' dataset, named COCO\_20k~\cite{DBLP:conf/eccv/VoPP20}.

\noindent
{\bf Evaluation metric.} Same to~\cite{DBLP:conf/eccv/VoPP20,DBLP:journals/pr/WeiZWSZ19,DBLP:conf/nips/VoSSPP21,DBLP:conf/bmvc/SimeoniPVRGBPMP21,wang2022tokencut}, we use the Correct Localization (CorLoc) metric for evaluation. In this metric, a predicted bounding box is considered correct if its intersection over union (IoU) score with one of the ground truth boxes in an image is greater than 0.5. 

\noindent
{\bf Implementation details.} 
We use the weights from the publicly available DINO pre-trained ViT model~\cite{DBLP:conf/iccv/CaronTMJMBJ21}. 
We adopt LOST/TokenCut as our basic unsupervised object detector $\mathcal{D}$, named \name-L/\name-TC.
For backbone architecture, we adopt ViT-S/16 and ViT-B/16~\cite{DBLP:conf/iclr/DosovitskiyB0WZ21}. We set $\sigma$=0.1 for \name-L and $\sigma$=1 for \name-TC. $\tau$ is set to $\sqrt{2}$ in our experiments. For multi-layer feature fusion, we only fuse the features from the last four layers.

\begin{figure*}[h]
	\centering
	\begin{tabular}{c@{\hskip 1.3pt}c@{\hskip 1.3pt}c@{\hskip 1.3pt}c}
		\includegraphics[width=0.21\textwidth, height=0.19\textwidth]{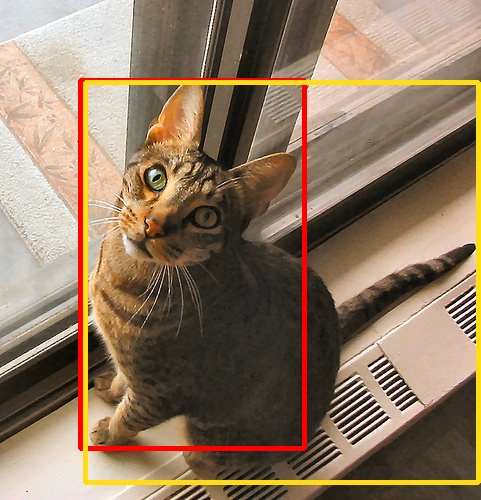} & 
		\includegraphics[width=0.21\textwidth, height=0.19\textwidth]{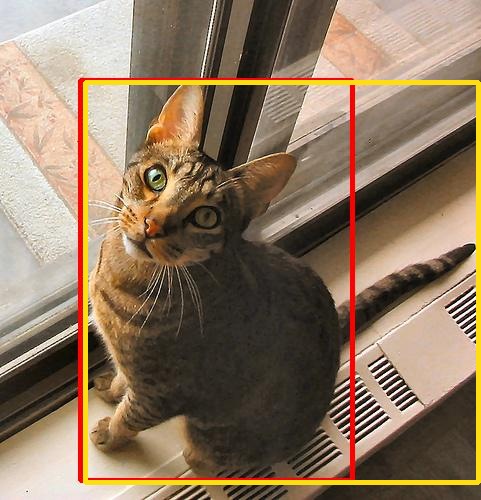} &
		\includegraphics[width=0.21\textwidth, height=0.19\textwidth]{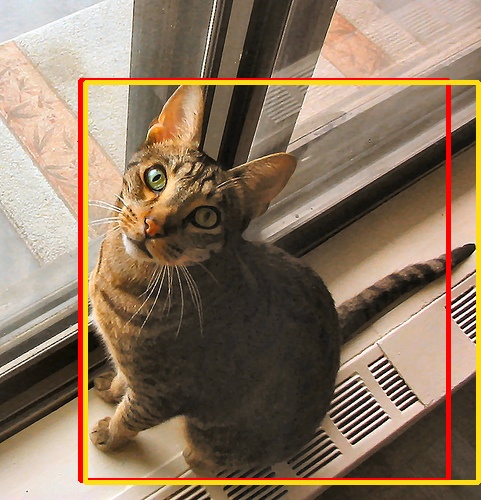} &
		\includegraphics[width=0.21\textwidth, height=0.19\textwidth]{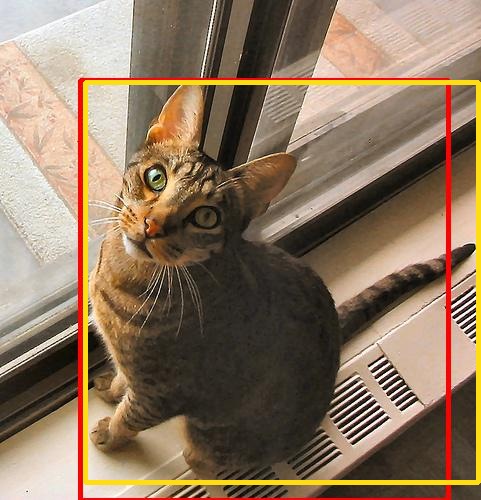} \\
		
		\includegraphics[width=0.21\textwidth, height=0.16\textwidth]{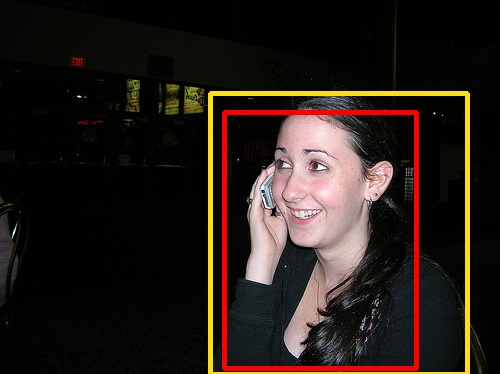} &
		\includegraphics[width=0.21\textwidth, height=0.16\textwidth]{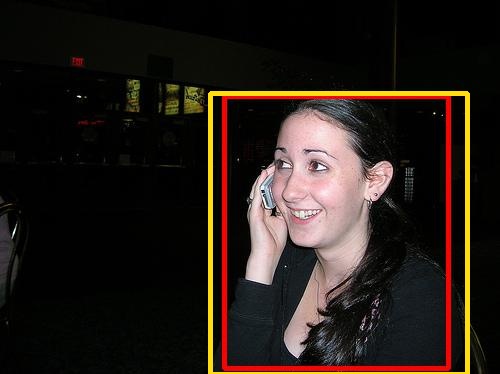} &
		\includegraphics[width=0.21\textwidth, height=0.16\textwidth]{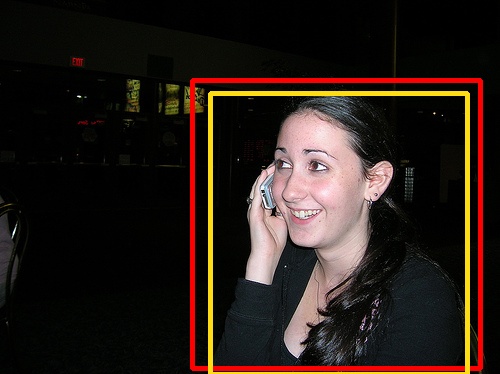} &
		\includegraphics[width=0.21\textwidth, height=0.16\textwidth]{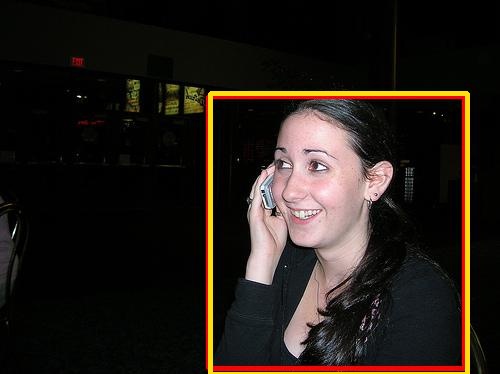} \\ 
		
		\includegraphics[width=0.21\textwidth, height=0.16\textwidth]{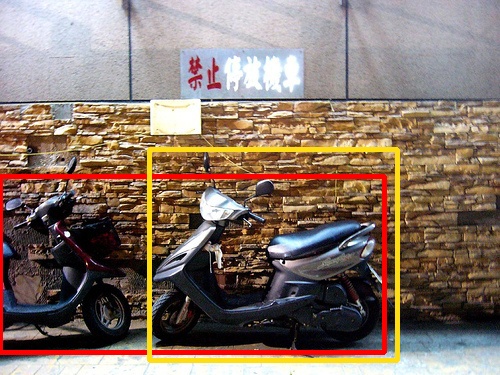} &
		\includegraphics[width=0.21\textwidth, height=0.16\textwidth]{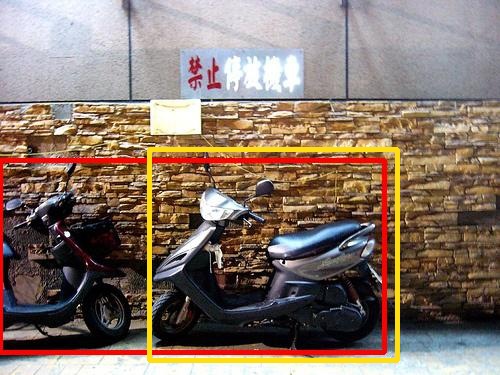} &
		\includegraphics[width=0.21\textwidth, height=0.16\textwidth]{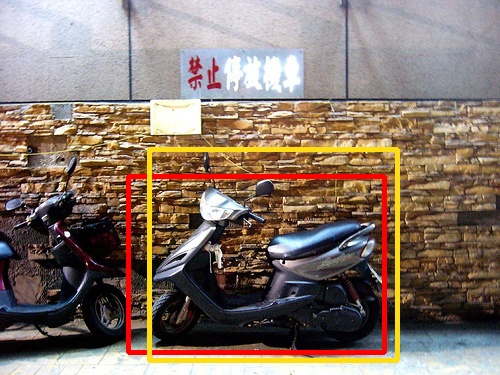} &
		\includegraphics[width=0.21\textwidth, height=0.16\textwidth]{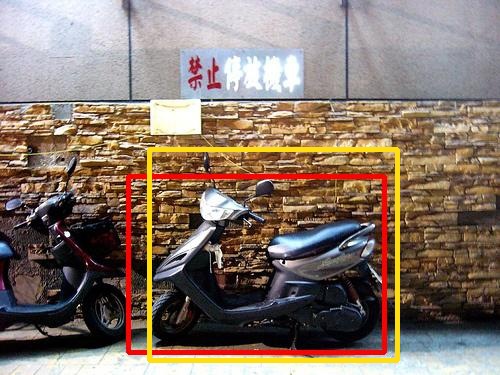} \\
		
		\includegraphics[width=0.21\textwidth, height=0.24\textwidth]{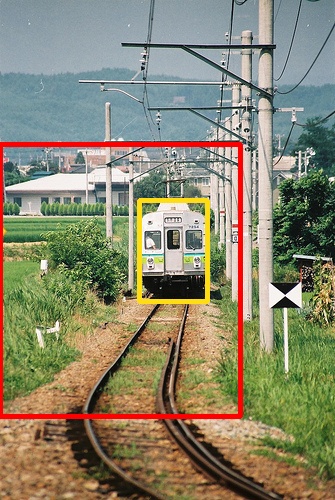} &
		\includegraphics[width=0.21\textwidth, height=0.24\textwidth]{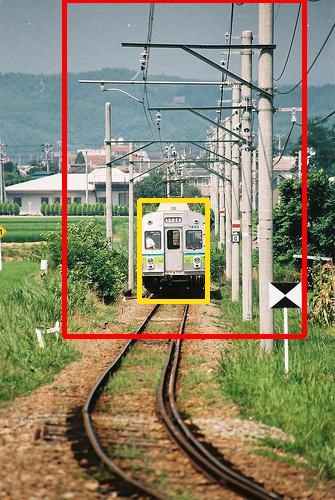} &
		\includegraphics[width=0.21\textwidth, height=0.24\textwidth]{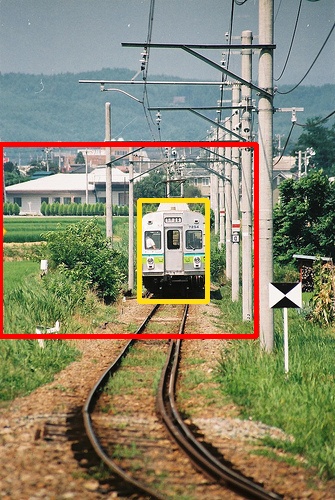} &
		\includegraphics[width=0.21\textwidth, height=0.24\textwidth]{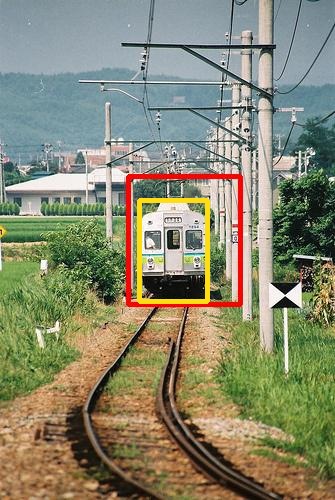} \\
		\makecell{(a) LOST} & \makecell{(b) TokenCut} & \makecell{(c) \name-L (ours)} & \makecell{(d) \name-TC (ours)} \\
	\end{tabular}
	\caption{{\bf Example results of unsupervised single-object discovery on VOC07 and VOC12.} In (a) and (b), we show the results obtained by LOST~\cite{DBLP:conf/bmvc/SimeoniPVRGBPMP21} and TokenCut~\cite{wang2022tokencut}. Our results are illustrated in (c) and (d). \textcolor{red}{Red} and \textcolor{Dandelion}{yellow} bounding boxes indicate the predicted bounding boxes and the ground-truth respectively. From top to bottom, the scale of the objects becomes smaller. 
	}
	\label{fig:disp}
\end{figure*}

\subsection{Main Results}
\label{sec:main_results}
\subsubsection{Unsupervised Single-object Discovery}
In Table \ref{tab:main_result}, we present unsupervised single-object discovery performance of \name~on three popular datasets. As shown in the table, our method outperforms the previous state-of-the-art, LOST and TokenCut, on all three datasets. Specifically, on VOC07, our method surpasses LOST and TokenCut by 2.5\% and 0.6\%, respectively. On VOC12, \name~obtains 67.7\% with ViT-S and 73.2\% with ViT-B, improving the performance of LOST and TokenCut by 3.7\% and 0.8\%, respectively. On the COCO\_20k dataset, \name~achieves 54.0\% and 59.7\%, which significantly surpasses LOST and TokenCut by 3.3\% and 0.7\%, respectively. In Fig. \ref{fig:disp}, we provide some visual results obtained by our method and the two baselines, \textit{i.e.}, LOST and TokenCut. It can be seen that our method boosts detection performance on various scales.

Additionally, following~\cite{DBLP:conf/bmvc/SimeoniPVRGBPMP21}, we also train a second-stage class-agnostic object detector (CAD) for unsupervised single-object discovery. Concretely, we train a Faster R-CNN~\cite{DBLP:conf/nips/RenHGS15} using the bounding boxes generated by \name~as pseudo-labels. It can be seen from Table \ref{tab:main_result} that our method outperforms the state-of-the-art by an average of 0.8\% on the three datasets. The consistent gain over baseline methods shows the effectiveness of \name~for unsupervised object discovery. 

\subsubsection{Unsupervised Object Detection}
Following LOST~\cite{DBLP:conf/bmvc/SimeoniPVRGBPMP21}, we also evaluate our method on unsupervised object detection. Similar to training a CAD, a Faster R-CNN detector is trained using pseudo-labels generated by our method. Concretely, to generate pseudo-labels for the class-aware detectors, we cluster the boxes produced by \name~via K-means algorithm and then map them to ground-truth classes via Hungarian matching~\cite{https://doi.org/10.1002/nav.3800020109} for evaluation. It is worth noting that no human supervision is involved during the training process. We use the \textit{Average Precision} metric with the threshold of IoU being 0.5, a common setting in PASCAL VOC detection. The results are shown in Table \ref{tab:od}. As we can see, we improve the state-of-the-art results of unsupervised object detection on VOC07 \verb'test' by a significant 4.1\%.

\subsection{Analysis and Discussion}
\label{sec:ablations}
In this section, we investigate the effectiveness of foreground guidance and multi-layer feature fusion modules. 
We conduct the following experiments with \name-L and use ViT-S/16 as the backbone.

\begin{figure*}[!ht]
	\centering
	\subfigure[]{
		\includegraphics[width=0.35\linewidth]{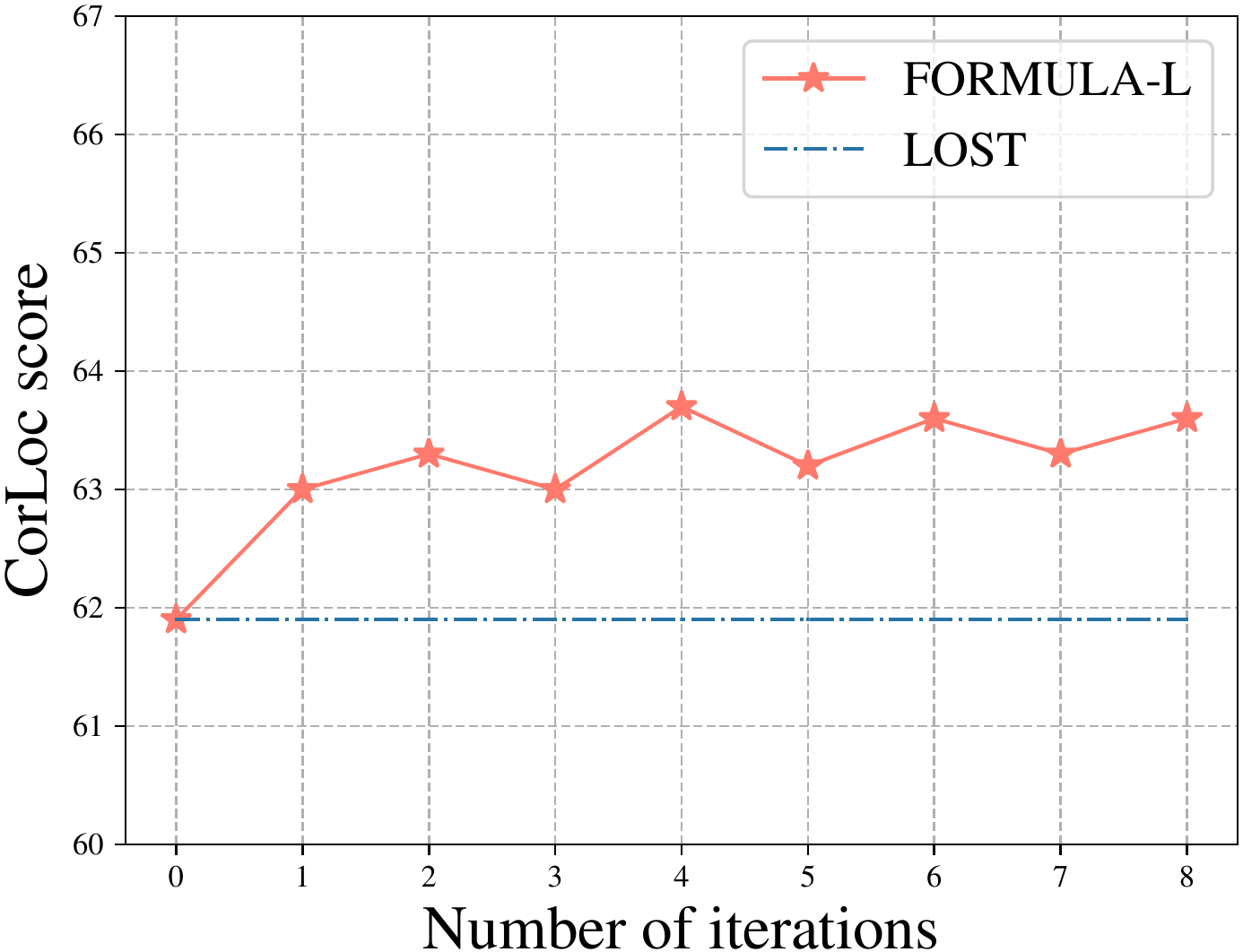}\label{fig:itera}
	}
	\subfigure[]{
		\includegraphics[width=0.35\linewidth]{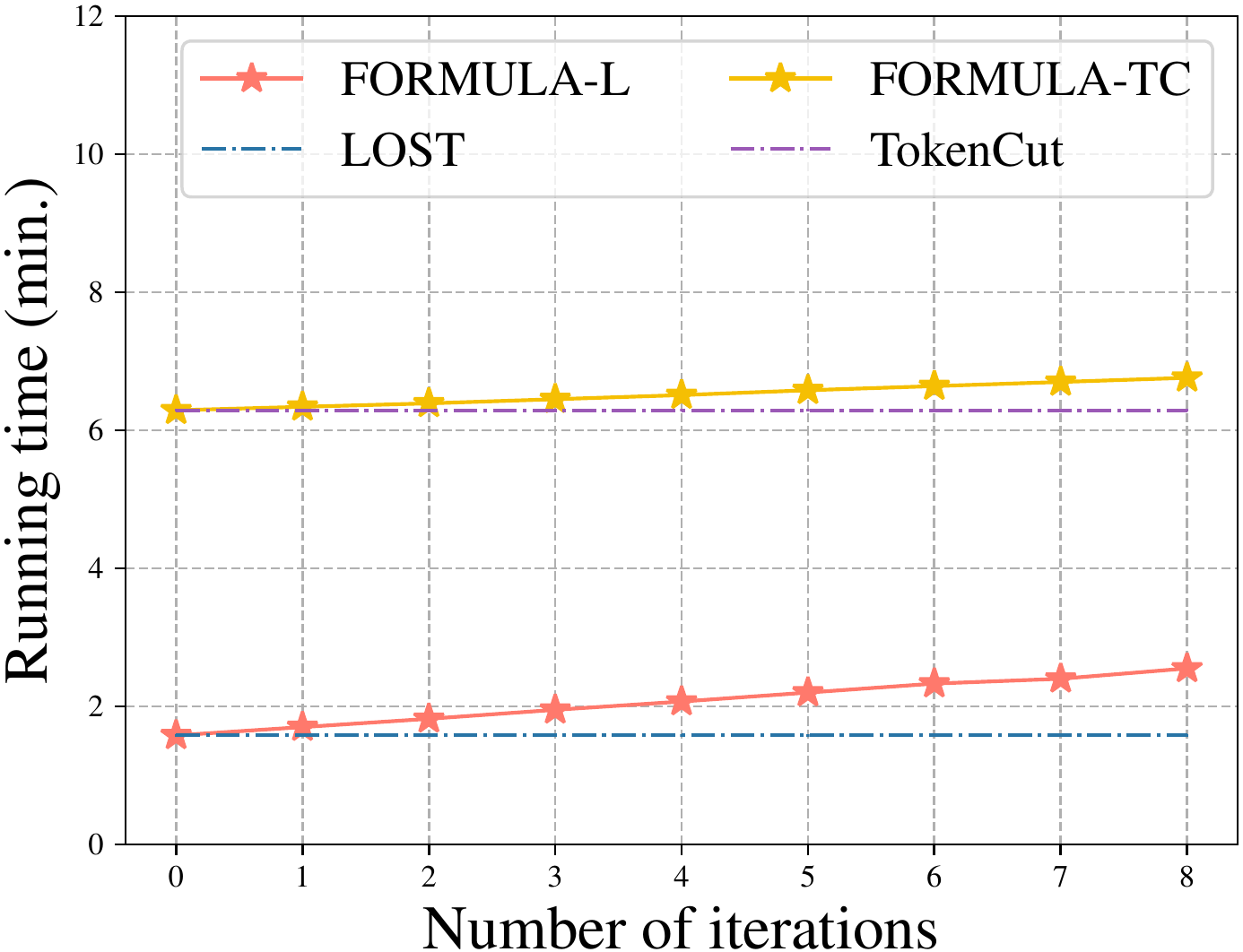}\label{fig:iterb}
	}
	\caption{{\bf Study of the iteration process.} (a) result variations with the number of iterations; (b) the running time of our method with different iteration numbers. Both experiments are conducted on VOC07 {\tt trainval}.}
	\label{fig:iter_study}
\end{figure*}

\begin{table}[htbp]
	\begin{center}
		\begin{tabular}{cc|c}
			\toprule
			\textit{Foreground guidance} & \textit{Multi-layer} & \textit{CorLoc$(\uparrow)$} \\
			\midrule
			& & 61.9 \\
			\checkmark & & 63.7 \\
			& \checkmark & 63.3 \\
			\checkmark & \checkmark & {\bf 64.4} \\
			\bottomrule
		\end{tabular}
	\end{center}
	\caption{{\bf Ablation experiments.} Results on VOC07 {\tt trainval}. ``\textit{Foreground guidance}'' and ``\textit{Multi-layer}'' represent foreground guidance module and multi-layer feature fusion module, respectively.}
	\label{tab:abl_modules}
\end{table}

\begin{table*}[htbp]
	\begin{center}
		\begin{tabular}{lcccc}
			\toprule Method & Backbone & VOC07$(\uparrow)$ & VOC12$(\uparrow)$ & COCO\_20k$(\uparrow)$ \\
			\midrule 
			LOST~\cite{DBLP:conf/bmvc/SimeoniPVRGBPMP21}& ViT-S/8~\cite{DBLP:conf/iccv/CaronTMJMBJ21,DBLP:conf/iclr/DosovitskiyB0WZ21} & 55.5 & 57.0 & 49.5 \\
			{\bf \name-L}  & ViT-S/8~\cite{DBLP:conf/iccv/CaronTMJMBJ21,DBLP:conf/iclr/DosovitskiyB0WZ21} & {\bf 57.9} & {\bf 61.3} & {\bf 49.6} \\
			\midrule
			LOST~\cite{DBLP:conf/bmvc/SimeoniPVRGBPMP21} &  ViT-S/16~\cite{DBLP:conf/iccv/CaronTMJMBJ21,DBLP:conf/iclr/DosovitskiyB0WZ21} & 61.9 & 64.0 & 50.7 \\
			{\bf \name-L} &   ViT-S/16~\cite{DBLP:conf/iccv/CaronTMJMBJ21,DBLP:conf/iclr/DosovitskiyB0WZ21} & {\bf 64.4} & {\bf 67.7} & {\bf 54.0} \\
			\midrule
			LOST~\cite{DBLP:conf/bmvc/SimeoniPVRGBPMP21}& ViT-B/16~\cite{DBLP:conf/iccv/CaronTMJMBJ21,DBLP:conf/iclr/DosovitskiyB0WZ21} & 60.1 & 63.3 & 50.0 \\
			{\bf \name-L}  & ViT-B/16~\cite{DBLP:conf/iccv/CaronTMJMBJ21,DBLP:conf/iclr/DosovitskiyB0WZ21} & {\bf 62.8} & {\bf 66.5} & {\bf 53.4} \\
			\bottomrule
		\end{tabular}
	\end{center}
	\caption{{\bf Analysis of the backbone.} CorLoc score comparison across three different backbones.}
	\label{tab:backbone}
\end{table*}

\subsubsection{Ablation Study}
We perform a set of ablation experiments to show the effectiveness of each component of our \name. Results of the ablations are shown in Table \ref{tab:abl_modules}. We can see that the foreground guidance module can bring an improvement of 1.8\% and the multi-layer feature fusion module improves model performance by 1.4\%. When both modules are applied, we have a total 2.5\% improvement. The results demonstrate the effectiveness of the two modules.

\subsubsection{Main Properties}
We discussed some of the properties of \name~here.

\begin{figure}[htbp]
	\centering
	\includegraphics[width=8cm]{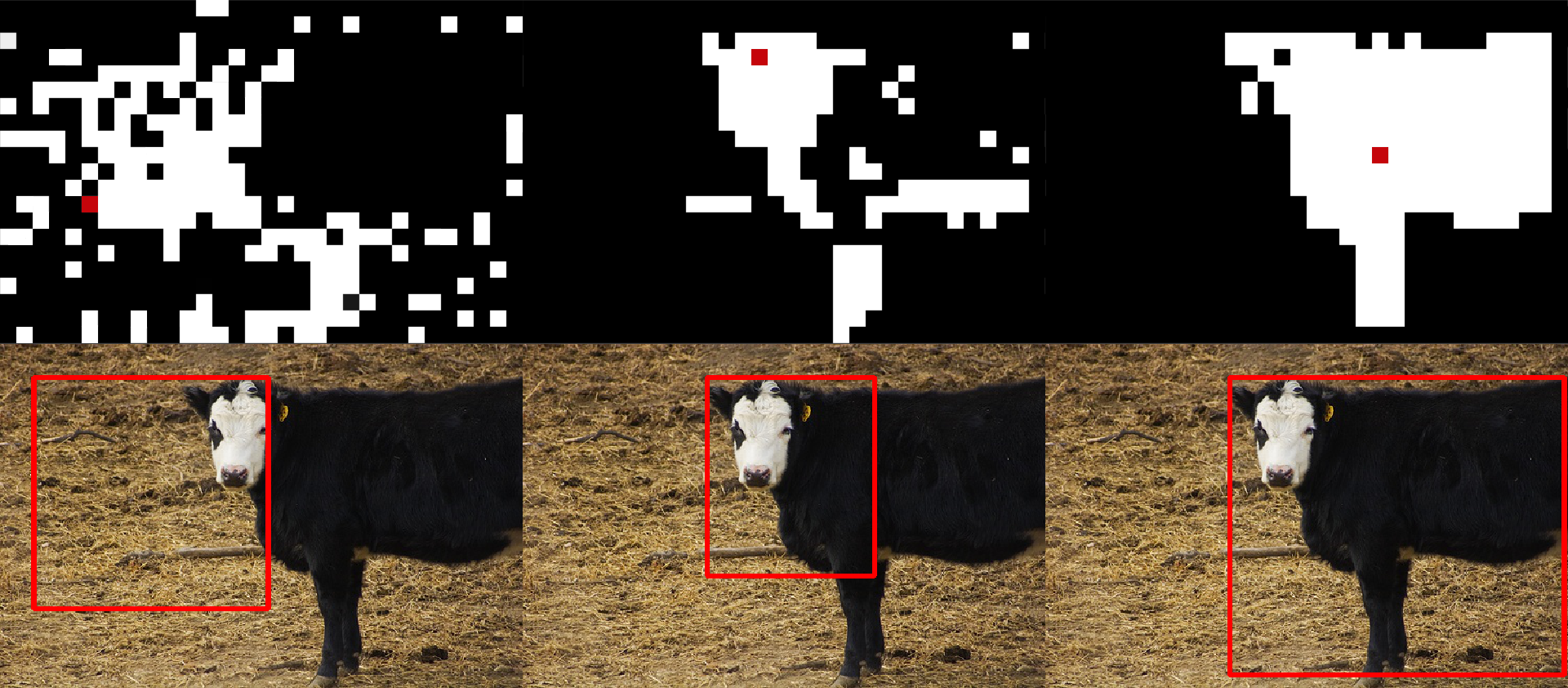}
	\caption{{\bf Study of foreground guidance.} We manually place the center $\bm{x}_{\mathcal{O}}$ (red points in the first row) in Eq. \ref{eq:eq3} at different locations. The center $\bm{x}_{\mathcal{O}}$ is placed outside the foreground object (left), at a corner part of the object (middle), and near the center of the object (right). The first and second rows are the resulting object masks and bounding boxes.}
	\label{fig:foreground}
\end{figure}

\noindent
{\bf Foreground guidance.} 
To better understand the effect of foreground guidance, we manually change the $\bm{x}_{\mathcal{O}}$ of probability map $P$ in Eq. \ref{eq:eq3} to different locations in the image. The effects are shown in Fig. \ref{fig:foreground}. When $\bm{x}_{\mathcal{O}}$ falls into the background, the model would miss the most area of the object. By contrast, if it is located inside the foreground object and closer to the center of the object, the model could better capture the semantic layout of the object and improve its performance. The result suggests that our guidance mechanism can help the model focus more on the region of interest.

\noindent
{\bf Backbone architecture.} 
In Table \ref{tab:backbone}, we show the results of \name with different Transformer backbones. 
We compare the results between ViT~\cite{DBLP:conf/iclr/DosovitskiyB0WZ21} small (ViT-S) and base (ViT-B) with a patch size of 8 or 16. It can be seen that \name~ consistently outperforms LOST across different backbone architectures, demonstrating the effectiveness and architecture generalizability of \name.

\begin{table}[!h]
	\begin{center}
		\begin{tabular}{c|ccc}
			\toprule
			$\sigma$ & VOC07$(\uparrow)$ & VOC12$(\uparrow)$ & COCO\_20k$(\uparrow)$ \\
			\midrule
			0.01 & {\bf 63.7} & {\bf 66.6} & {\bf 52.8} \\
			0.05 & {\bf 63.7} & {\bf 66.6} & {\bf 52.8} \\
			0.1 & {\bf 63.7} & {\bf 66.6} & {\bf 52.8} \\
			0.3 & 63.1 & 66.3 & 52.6 \\
			0.5 & 62.4 & 66.0 & 52.3 \\
			\bottomrule
		\end{tabular}
	\end{center}
	\caption{{\bf Ablation of $\sigma$.} CorLoc scores for different $\sigma$ on three datasets. The results are obtained using only the foreground guidance module.}
	\label{tab:sigma}
\end{table}

\noindent
{\bf Iteration number.} Fig. \ref{fig:itera} shows the influence of the number of iterations from 1 to 8. We find surprisingly that just one iteration can improve performance significantly.
An optimal value for our method is 4, which achieves the best performance of 63.7\%. More iterations may slightly reduce performance, which we assign to the random perturbations after convergence.

\noindent
{\bf Running Time.} We present the running time of our method for different numbers of iterations in Fig. \ref{fig:iterb}. Results of LOST and TokenCut are shown as well. We measure the inference time on all images of VOC07 \verb'trainval' with a single GTX TITAN X GPU. 
It can be seen that our method only brings marginal extra computational overhead even with 8 iterations.

\begin{table}[htbp]
	\begin{center}
		\begin{tabular}{lcc}
			\toprule
			Datasets & \name-L & \name-TC \\
			\cmidrule(lr){1-1} \cmidrule(lr){2-2} \cmidrule(lr){3-3}
			VOC07 & [2,1,1,6] & [3,5,1,1] \\
			VOC12 & [1,1,2,6] & [1,6,1,2] \\
			COCO\_20k & [0,2,3,5] & [2,7,0,1] \\
			\bottomrule
		\end{tabular}
	\end{center}
	\caption{{\bf Layer weights.} The layer weights are the relative weight of each layer, with the sum being one. The weight of the last layer is in front and the others are in the following order.}
	\label{tab:params}
\end{table}

\begin{table}[htbp]
	\begin{center}
		\begin{tabular}{cccc|c}
			\toprule
			$layer^{-1}$ & $layer^{-2}$ & $layer^{-3}$ & $layer^{-4}$ & \textit{CorLoc$(\uparrow)$} \\
			\midrule
			\checkmark & & & & 61.9 \\
			& \checkmark & & & 61.5 \\
			& & \checkmark & & 62.9 \\
			& & & \checkmark & {\bf 63.3} \\
			\bottomrule
		\end{tabular}
	\end{center}
	\caption{{\bf Ablation of the contribution of each layer.} We use features of each of the last four layers as input. `$layer^{-1}$' represents the last layer. The results are acquired on VOC07 {\tt trainval}.}
	\label{tab:abl_layers}
\end{table}


\begin{figure}[ht]
	\centering
	\subfigure[Input]{
		\includegraphics[height=1.8cm]{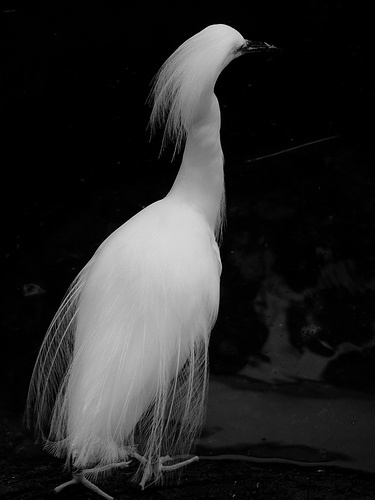}\label{fig_5a}
	}
	\subfigure[$layer^{-1}$]{
		\includegraphics[height=1.8cm]{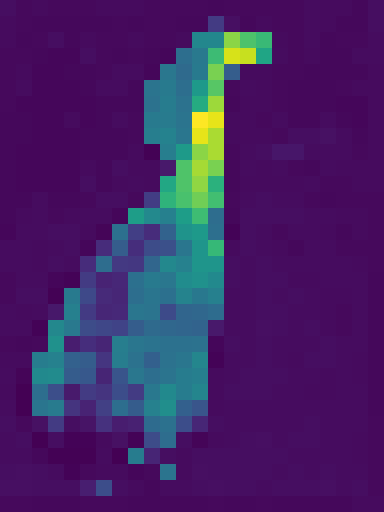}\label{fig_5b}
	}
	\subfigure[$layer^{-2}$]{
		\includegraphics[height=1.8cm]{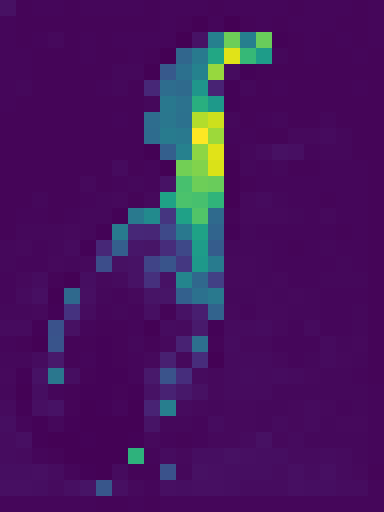}\label{fig_5c}
	}
	\subfigure[$layer^{-3}$]{
		\includegraphics[height=1.8cm]{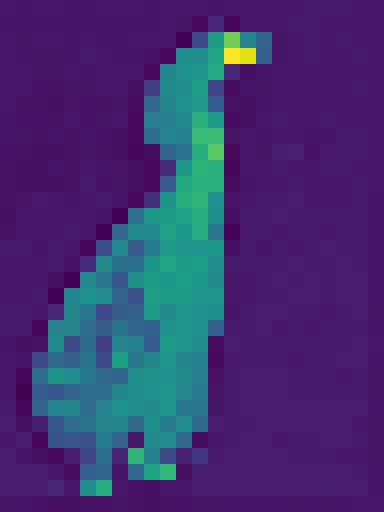}\label{fig_5d}
	}
	\subfigure[$layer^{-4}$]{
		\includegraphics[height=1.8cm]{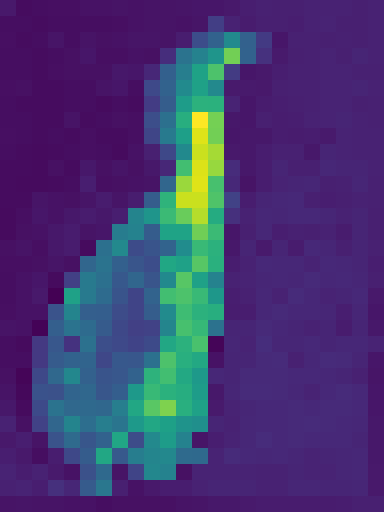}\label{fig_5e}
	}
	\caption{{\bf The activated areas of attention maps from the last four layers of ViT.} (a) Image sampled from VOC07~\cite{pascal-voc-2007}; (b)-(e) are the visualized attentions $F_{int}$.
	The deeper layers gather global information and focus on the discriminative parts of the object.
	}
	\label{fig:attn}
\end{figure}

\noindent
{\bf Analysis of $\sigma$.} We report the results of using different value of $\sigma$ in Table \ref{tab:sigma}. 
It can be observed that the performance slightly decreases as $\sigma$ increases above 0.1, which we attribute to the fact that, with a higher value of $\sigma$, the broader 2D Gaussian distribution would make the model focus on larger areas and thus hurt detection on small objects. 
Additionally, our method performs consistently with a $\sigma$ between 0.01 and 0.1. Thus, we simply adopt a typical value of 0.1 in our experiments.

\noindent
{\bf Multi-layer Fusion weights.} The weights of multi-layer fusion weights for our results in Table \ref{tab:main_result} are presented in Table \ref{tab:params}. \name~with LOST is better at capturing local information of the foreground, thus requiring more global knowledge from low layers to perform better. 
Different from \name-L, the information from the last two layers plays a more important role for \name-TC.

Besides, to better understand how each layer contributes to the overall performance, we conduct experiments using each of the four layers. The results are presented in Table \ref{tab:abl_layers}. We can see that low layers, such as the third and fourth ones, play a more important role in gathering features at different scales, which aligns with the results in Table \ref{tab:params}. We also visualize the attention maps of the last four layers in Fig. \ref{fig:attn}. 
The activated areas of the attention map vary at different scales for different layers.
These results together indicate the ViT features from different layers can help detect objects at various scales.

\section{Conclusion}
In this work, we propose \name, a simple and effective feature enhancement method for unsupervised object discovery. 
We utilize the foreground guidance acquired by an existing UOD detector to highlight the foreground regions on the feature maps and iteratively refine the segmentation predictions.
In addition, by fusing the multi-layer features from a self-supervised ViT, we further aggregate multi-scale information for the feature representation.
Our approach can work with any existing ViT-based unsupervised object discovery methods. Moreover, \name~achieves new state-of-the-art results on three datasets for unsupervised object discovery task. We hope our work could inspire more future research on enhancing ViT features for unsupervised visual learning.

\clearpage
{\small
	\bibliographystyle{ieee_fullname}
	\bibliography{egbib}
}

\end{document}